\documentclass[journal]{IEEEtran}
\ifCLASSINFOpdf \else \fi

\usepackage{graphicx}
\usepackage{comment}

\usepackage[cmex10]{amsmath}
\usepackage{amssymb}
\usepackage{color}

\usepackage{subfigure}

\usepackage{algorithmic}
\usepackage{array}
\ifCLASSOPTIONcompsoc
 \usepackage[caption=false,font=normalsize,labelfont=sf,textfont=sf]{subfig}
\else
 \usepackage[caption=false,font=footnotesize]{subfig}
\fi
\usepackage{url}
\usepackage[noadjust]{cite}

\begin{document}

\title{A Decentralized Cooperative Navigation Approach for Visual Homing Networks}

\author{{Mohamed Rahouti,~\IEEEmembership{Member,~IEEE,} Damian Lyons,~\IEEEmembership{Senior Member,~IEEE,} Senthil Kumar Jagatheesaperumal, and Kaiqi Xiong,~\IEEEmembership{Senior Member,~IEEE}}

\IEEEcompsocitemizethanks{\IEEEcompsocthanksitem M. Rahouti is with the Department of Computer \& Information Science, Fordham University, NY, NY USA (email: mrahouti@fordham.edu).
\IEEEcompsocthanksitem D. Lyons is with the Department of Computer \& Information Science, Fordham University, Bronx, NY USA (email: dlyons@fordham.edu).
\IEEEcompsocthanksitem S.K. Jagatheesaperumal is with the Department of Electronics \& Communication Engineering, Mepco Schlenk Engineering College, Sivakasi, Tamil Nadu, India (e-mail: senthilkumarj@mepcoeng.ac.in).
\IEEEcompsocthanksitem K. Xiong is with Cyber Florida and ICNS Lab, University of South Florida, Tampa, FL USA (email: xiongk@usf.edu).
}
\thanks{Manuscript received XX, 2023}}

\markboth{IEEE IT Professional,~Vol.~X, No.~X, XX~X}%
{Rahouti \MakeLowercase{\textit{et al.}}: Blockchain-Enabled Cooperative Navigation Strategy For Visual Homing Networks}

\maketitle

\begin{abstract}
Visual homing is a lightweight approach to visual navigation. Given the stored information of an initial `home' location, the navigation task back to this location is achieved from any other location by comparing the stored home information to the current image and extracting a motion vector. A challenge that constrains the applicability of visual homing is that the home location must be within the robot's field of view to initiate the homing process. Thus, we propose a blockchain approach to visual navigation for a heterogeneous robot team over a wide area of visual navigation. Because it does not require map data structures, the approach is useful for robot platforms with a small computational footprint, and because it leverages current visual information, it supports a resilient and adaptive path selection. Further, we present a lightweight Proof-of-Work (PoW) mechanism for reaching consensus in the untrustworthy visual homing network.

\end{abstract}


\IEEEpeerreviewmaketitle

\section{Introduction} \label{sec:intro}
Visual homing is a navigation strategy employed often in mobile robots in which a motion vector towards a target destination is calculated by comparing the image captured at a particular instant with that of the destination. For real-time autonomous navigation of mobile robots, and based on industrial demands, visual homing is gaining more popularity with its reliable characteristics inspired by biological navigation.

\textcolor{black}{Several approaches exist for visual homing calculations \cite{lee2018visual}: holistic methods transform whole images, while feature-based methods, like the average landmark vector (ALV), extract and compare features. The ALV requires compass information, but correspondence-based methods don't, utilizing instead relative motion vectors and potentially SIFT scale \cite{ji2018three} or stereo information for depth cues. While visual homing aids in local-area navigation when combined with obstacle avoidance \cite{Fu_2018}, a challenge is that the robot must initially spot the destination in its field of view (FOV). Often, a panoramic or wide FOV system is used. We suggest using FOVs from multiple robots in a team to address this, as a target might be in one robot's FOV even if another can't see it \cite{Lyons_2022}.}

In such a scenario, the main challenge is how to establish trustworthy communication for driving the autonomous navigation of robots. First, in a visual homing network, the target position can be dynamic and needs to be communicated to the robots involved in the cooperative navigation. Second, trustworthy communication infrastructures are needed for establishing sustainable links among the visual homing network. The trustworthiness is ensured as each data entry in the ledger is cryptographically linked to the previous one, forming a chain of blocks. This ensures that any unauthorized modifications or tampering attempts can be detected, promoting data integrity and authenticity.

\begin{figure*}[!t]
\centering
\centerline{\includegraphics[height=15cm, width=16cm]{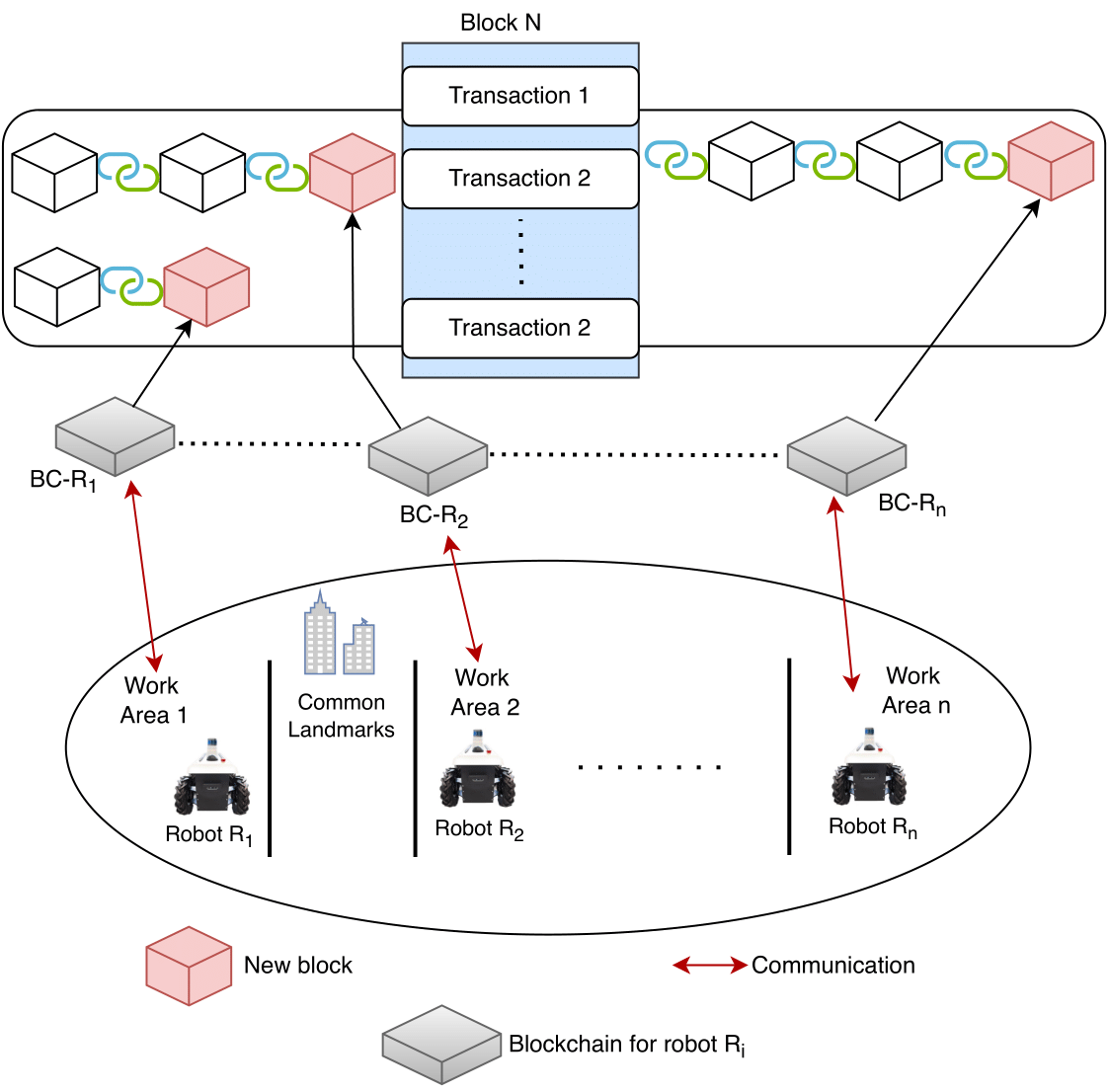}}
\caption{Architectural design of the blockchain-enabled visual homing network.}
\label{fig:usecases}
\end{figure*}

This paper addresses the limitations of visual homing-enabled robotic environments by proposing a novel blockchain-enabled visual homing (BC-VH) framework that leverages decentralized blockchain technology for lightweight navigation and FOV sharing. This study explores challenges related to efficient navigation and resource consumption, presents the BC-VH architectural design, and discusses its implementation, use case scenario, consensus mechanism, and security assessment. The main components of the proposed architecture are depicted in Fig. \ref{fig:usecases}. A lightweight consensus mechanism is further deployed to enable timely verification and validation of FOV updates. This mechanism enables visual homing team robots to achieve consensus (agreement) in a cooperative manner rather than competitive. The article addresses the challenges in visual homing networks with the idea of BC-VH solutions focused on the following contexts.

\begin{itemize}
    \item An architecture to enhance visual homing environments: The proposed architecture will allow individual team robots to efficiently share and identify up-to-date common landmarks at a low operational cost and in a timely manner.
    \item A consensus suitable for resource-constrained robot navigation systems: Creating a blockchain typically requires significant computation power, which thus depletes the computational resources of individual robots in visual homing environments. The proposed consensus mechanism is lightweight, and team robots share the same access control list (ACL) secured by the blockchain. Therefore, burdens on the computational complexity are alleviated for individual robots.
\end{itemize}

In this article, we first present the core aspects of visual homing and highlight relevant research from the literature. Next, we provide a blockchain framework to establish trustworthy decentralized communication for visual homing. Further, we describe use case scenarios, configurations, and metrics for the evaluation of BC-VH. We also show how the proposed consensus mechanism can be incorporated into BC-VH for its assessment.

\section{Background} \label{sec:background}

Visual homing as a robotic navigation technique arguably originated in computer models with the inspiration from insect navigation in 1983. The FOV for an insect can be very wide, even 360$^0$, and for this reason, a panoramic camera is also often used for robot visual homing \cite{lyons2020evaluation}. 

\subsection{Visual Homing}
The process of visual homing implementation involves image acquisition, feature extraction, and image warping to transform and align the acquired image with a reference frame. By comparing the warped image with the reference, the robot's precise location is determined, allowing for effective navigation and control toward the desired destination. Approaches to solving the visual homing problem can be separated into whether they consider each image as a whole (holistic) or whether information or features are extracted from each image and compared to each other (feature-based comparison). A remarkable approach that can be used is warping, a holistic approach where the current image is warped according to different movement vectors, and an optimal movement vector is selected based on the warped image most similar to the home image. Image distance methods rely on observing that image distance measures will increase proportionally to the spatial distance between locations at which the images have been taken. A matched-filter image distance algorithm can be used for visual homing. Further, in~\cite{ji2022adaptive}, the authors focused on using stereo-depth information for adaptive correction of landmarks to perform real-time homing tasks.

\subsection{Wide Area (Scalable) Visual Homing}
The restriction that the home location is in the initial FOV for homing to begin has been addressed by a number of researchers. A traditional method is based on ant navigation in which the robot stores images as it moves in the environment. Navigation to a location that is not in view is accomplished by reprising visual homing to the sequence of images recorded on a previous trip to the distant location. Stelzer et al. \cite{stelzer2018towards} addressed this problem for `scalable' visual homing using an algorithm based on bearing angles, where they proposed a data structure that stores and efficiently leverages the intermediate imagery as configurations of landmark bearings called view frames. Image sequence approaches to wide-area visual navigation require that the robot explores the work area, and determines when to store a new intermediate image by inspecting the distance traveled, and estimating the landmark dissimilarity \cite{stelzer2018towards}. All of these aforementioned approaches construct a map data structure representing the motion and observations of the robot over the working area.

The wide area visual navigation (WAVN) \cite{Lyons_2022} approach addresses the problem of visual homing to a distant (out of sight) location in a different way. Prior approaches carry forward the original insect inspiration for visual homing, for instance, a single insect wandering in a landscape. WAVN starts with a team of robots distributed over a wide area and in wireless communication with each other. If a team member is driven by a task to go to a location that is not within its FOV, then it can leverage the fact that the location may be visible or have recently been visible to one of the other team members (or indeed fixed camera assets, as may exist in an urban area).

If team member $A$ needs to travel to a specific home location, but that location is not in view, then robot $A$ tries to find another robot $B$ that can see the home location. If robot $B$ can see the home location, then $A$ and $B$ must next identify a common visual landmark that can be used as an intermediate visual target that $A$ can use to travel to the vicinity of $B$ and from there to the final destination. While this example uses only two robots and one shared landmark, the approach can be generalized to a sequence of intermediate landmarks. Lyons and Petzinger \cite{Lyons_2022} evaluated approaches to identifying these common landmarks, proposing a combination of deep-learning and SIFT-based techniques that significantly outperform a purely feature-based approach. The WAVN approach does not require a map data structure and it is hence useful for implementation on small or computation-constrained platforms and in certain situations where GPS is not (readily or consistently) available. The visual information in WAVN is always up to date, reflecting the current visibility of the robot team. However, it is appropriate for long-term deployments in areas whose appearance may change with weather or seasons and so forth and also in which maps might easily become out of date. Furthermore, the communication of visual information among the team of robots is a crucial part of this approach, which raises security concerns.

\subsection{Security of Imaging}

Challenging scenarios do occur during cooperative perception in a visual homing network. Malicious nodes could send false information about image cues to the robot team members and thereby influence or alter their navigation policies. During the navigation, different security factors need to be commonly considered, where the unpredictable behavior of the fake robotic nodes in the visual homing networks could be dealt with ease. Apart from the commercial off-the-shelf security frameworks, the launch of blockchain has created opportunities for visual homing networks in framing cooperative navigation strategies. Additionally, blockchain frameworks are typically being deployed in robotic applications for accurate prediction of false information from unauthenticated nodes and safeguarding them against deception. With the communication among the robots in the visual homing network, being established through the blockchain framework, tamper-proof records of information are maintained in the network and the robot helps to identify the inconsistencies. Depending on the scenarios, every single robotic node in the work area will be registered in the blockchain network. Obviously, with the increase of the nodes in the network, a smart contract and consensus management need to be strengthened for secure communication and framing trustworthy navigation strategies.

The integration of features from multiple images streamed from various nodes of visual homing networks also poses a great threat in being potentially exploitable by unauthorized nodes. Introducing blockchain frameworks could discover the related features and focus to identify the risky feature extracts or tampered features from the malicious nodes in the visual homing network. To enable non-trivial feature extraction and trustworthy communication among the nodes in the visual homing network, the blockchain framework can harness the feature extraction process by introducing secure ledger transactions. A block in the network ensures continuous end-to-end connectivity through smart contracts established on a feature extraction process in the network.

\section{State of the Art} \label{sec:related}
Several research studies have integrated blockchain technology into a wide range of applications, including, but not limited to, the Internet of Things (IoT), smart cities, connected autonomous vehicles (CAVs), and robotic systems. The integration of such a decentralized technology in robotic environments, such as visual homing, can help tackle the security challenges and limitations beyond distributed/decentralized decision-making for robotic navigation tasks. Castello et al. \cite{castello2018blockchain} presented a blockchain-enabled framework to improve the decision-making and communication security in swarm systems, while Fernandes and Alexandre \cite{fernandes2019robotchain} proposed an event management framework by leveraging the decentralized blockchain and Tezos technology. Strobel et al. \cite{strobel2018managing} further developed blockchain-based collaborative decision-making in byzantine robots using smart contact coordination strategies.

Empowering blockchain characteristics combine the advantages of security and energy efficiency for the practical implementation of the swarm of UAVs together~\cite{alsamhi2022blockchain}. They involve multi-drone connectivity and collaboration to enhance security and achieve consensus for exploring their navigating environment. Although the means to address the lifecycle and trust management of blockchain might be still short, the flexibility of the UAVs in joining the network, gathering quality data, and preventing assaults can address such issues by hovering across the target area. Further, it also expands the benefits of enhancing the energy efficiency and connectivity of a navigating environment.

When exploring the ways to formally deploy blockchain-based trust for multi-robot collaborative tasks \cite{moron2022uwb}, it turns out that the remarkable efforts originate from the edge knowledge inference model. In particular, we can formulate an edge inference process in terms of the collaborative knowledge-based blockchain consensus strategy proposed by Li et al.~\cite{li2021blockchain} in the contest establishing collaborative tasks among multi-robotic systems. To do so, an emergency rescue application is chosen as a case study to evaluate the framework that measures the accuracy and latency of the edge knowledge inference.

Distributed landmarks can be created and deployed for establishing VH to enable strategic navigation for mobile robots based on requests~\cite{sun2022novel}. By considering equal distance assumptions on the chosen landmarks, the VH strategy can be implemented based on the pseudo-isometric distribution vector to offer adaptive navigation and alleviate environmental challenges. The unevenness and randomness issues in the distribution of landmarks can be addressed by the vector pre-assign mechanism based on pseudo-isometric characteristics. Such a VH strategy can enable better precision and be capable of maintaining the computation speed with the efficient implementation of multiple robots in the scenario.

Blockchain technology can be used as a communication tool within multirobot systems, for leaders to broadcast directions to the whole group \cite{ferrer2021following}. Distributed ledger and cryptocurrency platforms, such as IOTA, can also be leveraged along with Robot Operating System (ROS) to provide scalable robotic systems and network partition tolerance \cite{keramat2023partition}. Additionally, blockchain can be employed for certificate verification purposes using robotic technologies \cite{malsa2021framework}.

Although existing state-of-art solutions enable specialized robotic teams to carry out individual-specific tasks (e.g., navigation, flocking, etc.), very little to no work considers leveraging a decentralized technology with visual homing-based robotic platforms. Given the trustworthiness property, low operational cost, efficient access control, and provenance, the proposed BC-VH solution will not only enhance navigation tasks in a visual homing environment but also enable further smart city-based robot-driven applications and use cases. The proposed BC-VH solution allows the individual team robots to efficiently and reliably share and identify up-to-date common landmarks in a timely, trustworthy, and secure manner, with a low overhead/operational cost \cite{aditya2021survey}.

\section{The Design Principle of BC-VH} \label{sec:design}

In this section, we propose a novel architecture based on blockchain and its associated design principles involved with consensus and smart contract management such that the proposed architecture has the potential to address some of the aforementioned challenges in the visual homing network. It is worth noting that communication between team robots is assumed secure in a locally controlled and trusted network environment. This is primarily due to the limited access and trust in the local environment. Specifically, access to the network is typically limited to authorized devices within the defined physical area (i.e., this assumption is based on the understanding that only trusted team robots and devices are connected to the controlled network). \textcolor{black}{A manufacturing plant, for instance, is expected to have stringent access controls, both physically (entry and exit points) and digitally (network access). Thus, the assumption of secure communication in locally-controlled robotic network arises from the inherent controlled nature of the environment, the isolation from external networks, and the rigorous access controls, both physical and digital.}

\subsection{Architectural Design of BC-VH}

Fig. \ref{fig:usecases} depicts the architectural design of our proposed BC-VH prototype based on the conventional workflow of blockchain technology. The blockchain network can be simulated using tools like Hyperledger Fabric or Ganache, which is a local Ethereum blockchain for development and testing purposes. In cases where resource-constrained devices are involved, alternative blockchain technologies must be designed to be lightweight and optimized for constrained visual homing environments. Thus, we propose  a distributed ledger framework that eliminates the need for traditional blocks and miners, resulting in a lightweight and scalable solution for resource-constrained robotic systems. A FOV represents a set of six panoramic images, and it efficiently fits into a block transaction as an ordinary data field.

In Fig. \ref{fig:usecases}, $BC-R_s$ generates the block hash (BH) for the next block to include valid recent panoramic views of individual team robots. The established block is then broadcasted to all $BC-R_s$ inside the visual homing environment for verification. Once the new block is successfully validated upon reaching a network consensus, it will be appended to the local ledger of each team robot. 

\subsection{Use Case and BC-VH Working Scenario}
To prove the concept discussed in the aforementioned sections, we consider a case study derived the inspiration from the existing deployment of visual homing, for which we have included a blockchain framework for the trustworthy cooperative navigation of robots.

\begin{figure}[htbp]
\begin{tabular}{ccc}
\includegraphics[width=0.4\columnwidth]{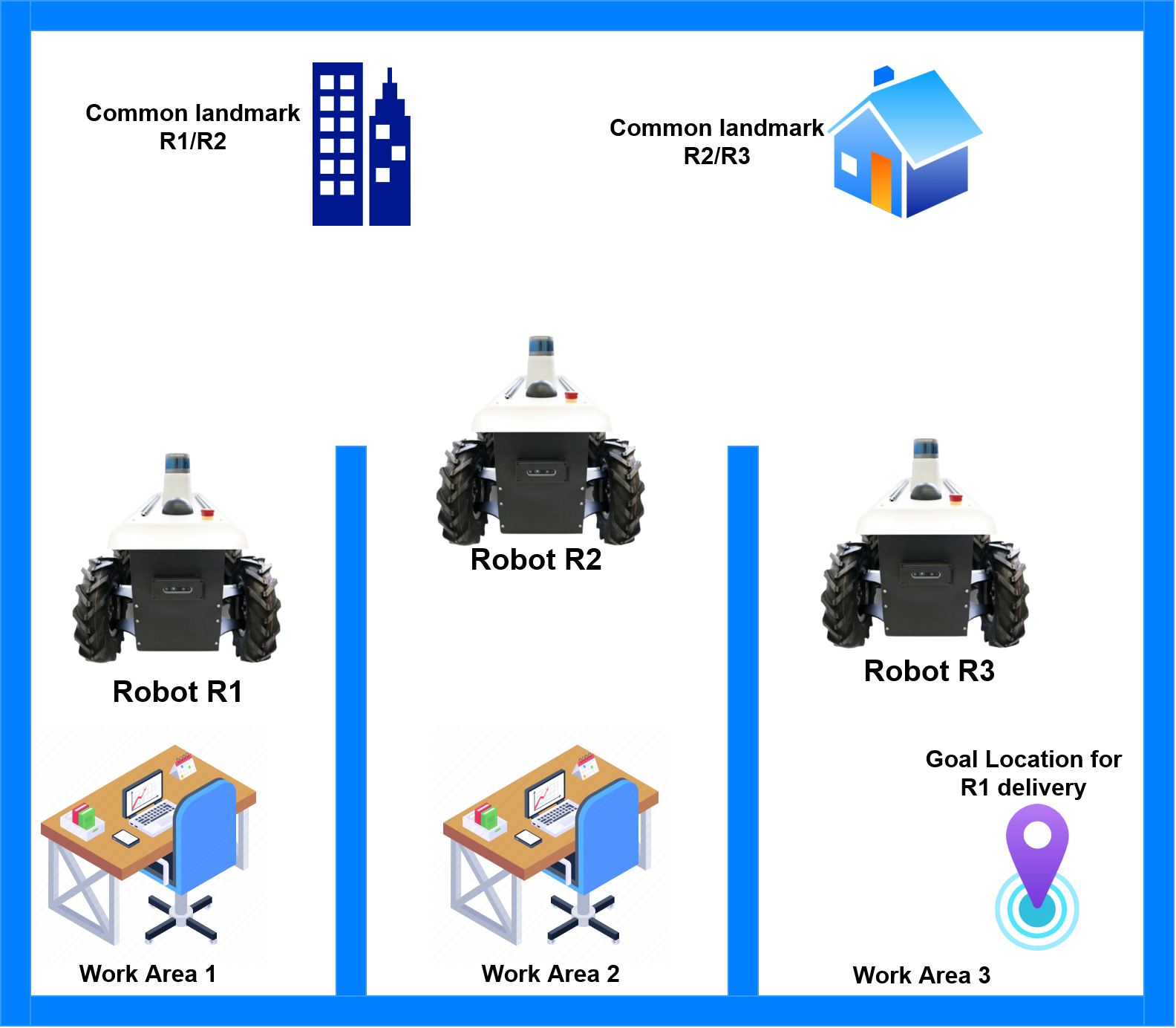}(a)&
\includegraphics[width=0.4\columnwidth]{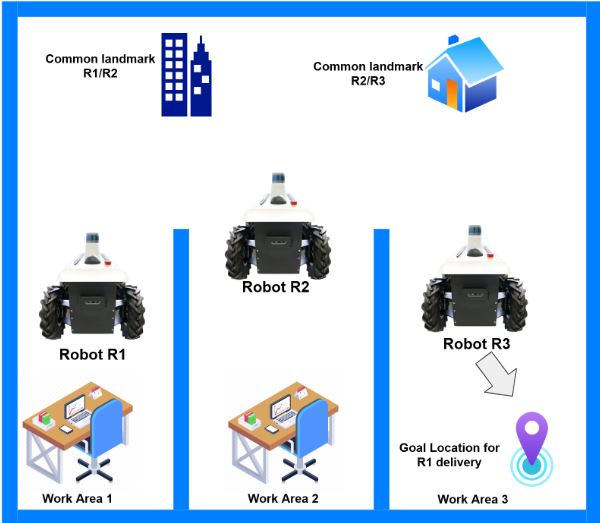}(b)\\
\includegraphics[width=0.4\columnwidth]{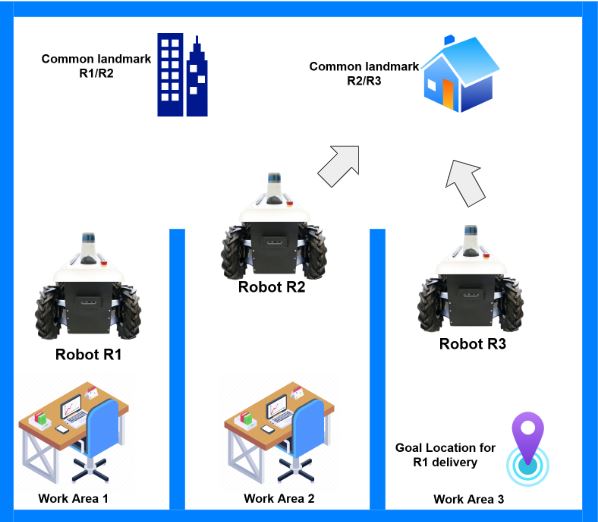}(c)&
\includegraphics[width=0.4\columnwidth]{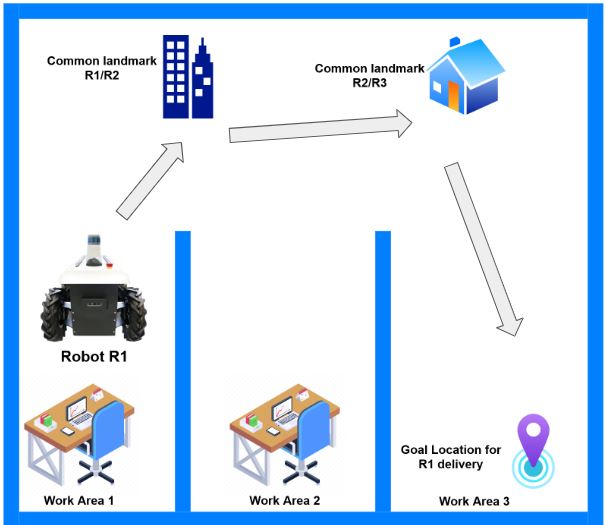}(d)
\end{tabular}
\caption{Navigation use case in a small-scale visual homing environment.}
\label{fig:usecase}
\end{figure}

In a small-scale visual homing environment with three team robots, $R_1$, $R_2$, and $R_3$, the blockchain-supported navigation scenario represented in Fig. \ref{fig:usecase} is established as follows:

\begin{figure*}[!t]
\centering
\centerline{\includegraphics[height=15cm, width=16cm]{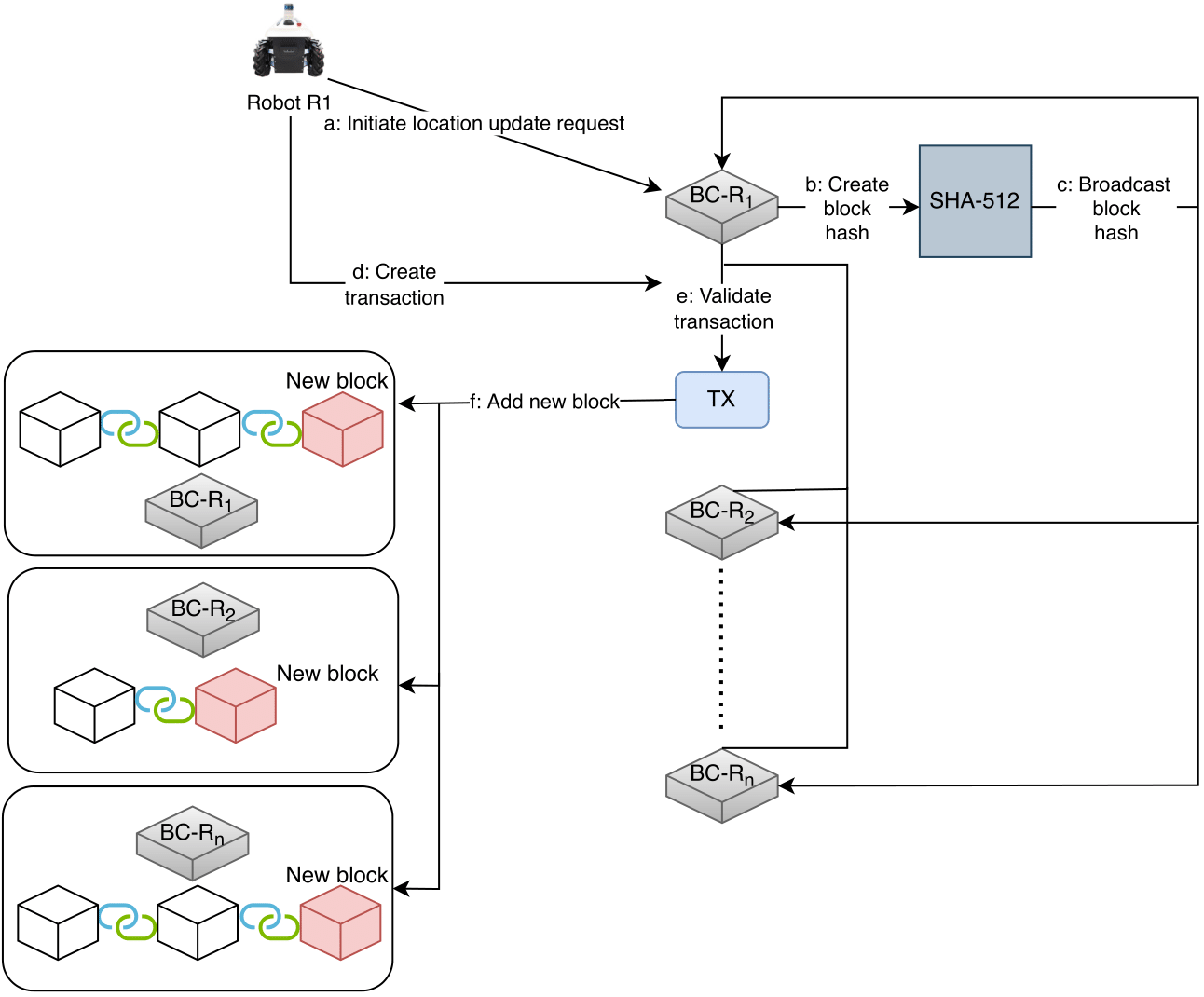}}
\caption{Visual homing consensus establishment in BC-VH.}
\label{fig:consensus}
\end{figure*}

\begin{enumerate}
    \item If Robot $R_1$ is unable to see the goal directly, it will first check which robot ($R_2$ or $R_3$ or both) is able to see the target location. In the scenario given in Fig. \ref{fig:usecase}(b), robot $R_3$ can see the target location. It is checked by the robot $R_1$ by looking in the transaction record ($BC-R_1$ ledger) of the last block (i.e., the most up-to-date FOVs) and $R_1$ navigates to the target location.
    \item $R_1$ next checks if that robot has a common landmark with $R_2$ or $R_1$ by checking the ledger. In the presented scenario, there is one common landmark between robots $R_3$ and $R_2$, Fig. \ref{fig:usecase}(b).
    \item $R_1$ next checks if $R_2$ and $R_1$ have a common landmark by checking the ledger. In this case, one common landmark is found.
    \item Finally, robot $R_1$ can now know the landmark path to the desired destination and complete the navigation, Fig. \ref{fig:usecase}(d).
\end{enumerate}

\section{Consensus Mechanism in BC-VH} \label{sec:consensus}
The consensus mechanism represents the core of blockchain technology, enabling a decentralized verification and validation of transactions. Specifically, PoW is deployed in the proposed prototype as the consensus mechanism to enable fairness among the team robots over the decentralized environment \cite{strobel2018managing}. As there are memory and energy constraints in visual homing robots, energy consumption, memory utilization, and convergence time are key concerns to consider when one designs a consensus mechanism in such a system. Therefore, we propose using a lightweight, enhanced PoW to provide a decentralized consensus mechanism for validating transactions and maintaining the integrity of the blockchain network. Such a lightweight consensus can also be suitable for resource-constrained platforms such as fog and cloud environments.

In this consensus mechanism, the process of reaching an agreement (consensus) by the team robots is depicted in Fig. \ref{fig:consensus}. The transaction representing the FOV update for a particular robot $R_i$ will be broadcast to the corresponding $BC-R_i$ for verification before it is broadcast to all $BC-R_s$ of other team robots for validation. Further, $SHA-512$ is the designated hash function used for hashing computation. Once the $BC-R_s$ ($BC-R_1$, $BC-R_2$, and $BC-R_3$) receive the broadcast transaction, they will cooperatively operate to compute the block hash and update their corresponding ledger.

It is important to note that the proposed application of blockchain to visual homing is an adaptation of specific aspects of PoW rather than the complete consensus mechanism itself. It aims to address some challenges (e.g., integrity, trustworthiness, and decentralized data sharing) in lightweight robot communication and navigation tasks by leveraging the key concepts from PoW, including block verification/validation and forking resolution. This adaptation's actual implementation and effectiveness may also depend on the specific requirements and characteristics of another robot system in question. The proposed blockchain-assisted visual homing can be feasibly reproduced on physical/real-world navigation robots in large-scale, GPS-denied environments \cite{lyons2023wavn}.

\section{Preliminary Performance and Security Assessment}

\begin{figure}[htbp]
\centering
\centerline{\includegraphics[height=4.5cm, width=8.5cm]{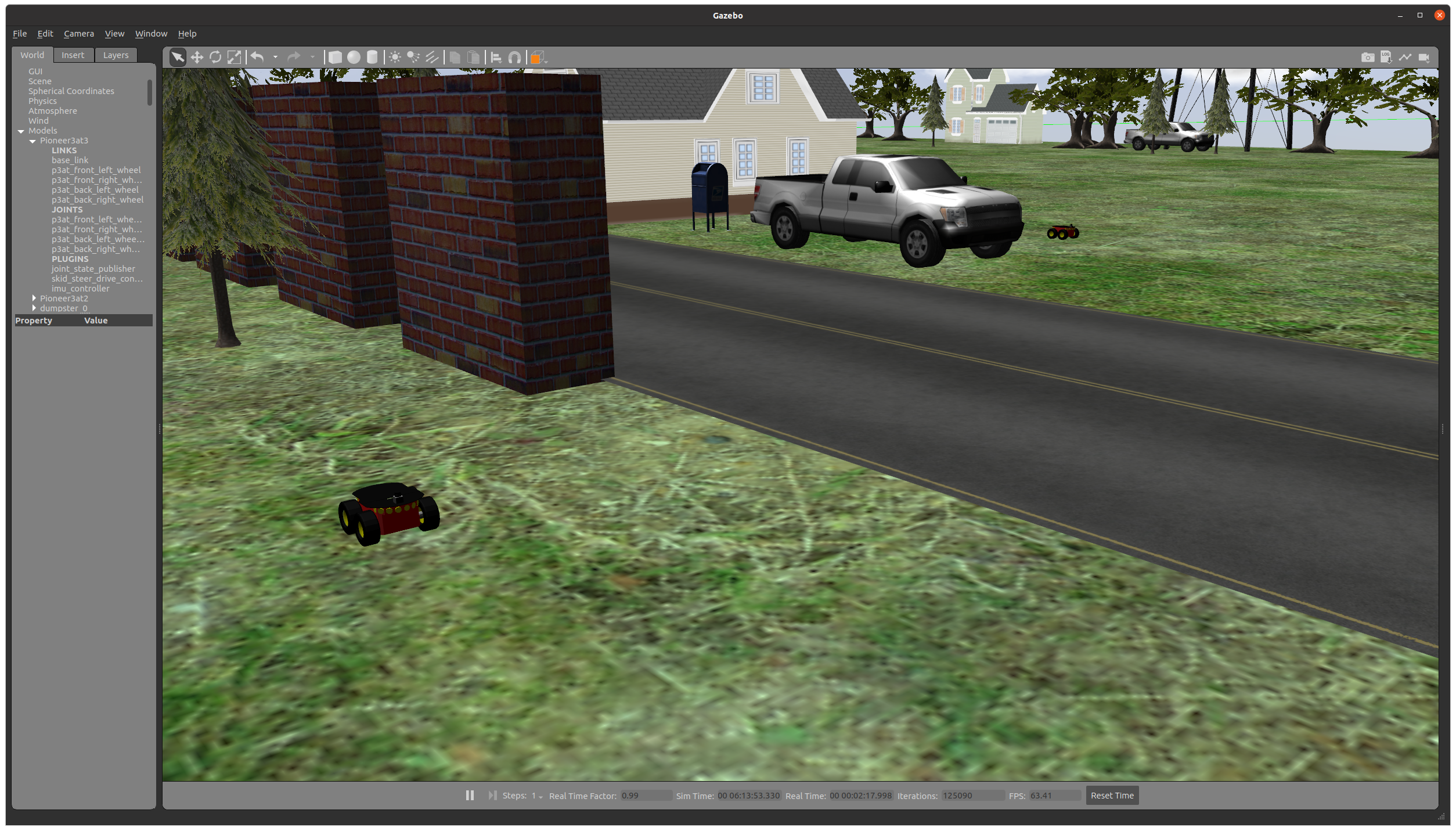}}
\centering
\centerline{\includegraphics[height=4.5cm, width=8.5cm]{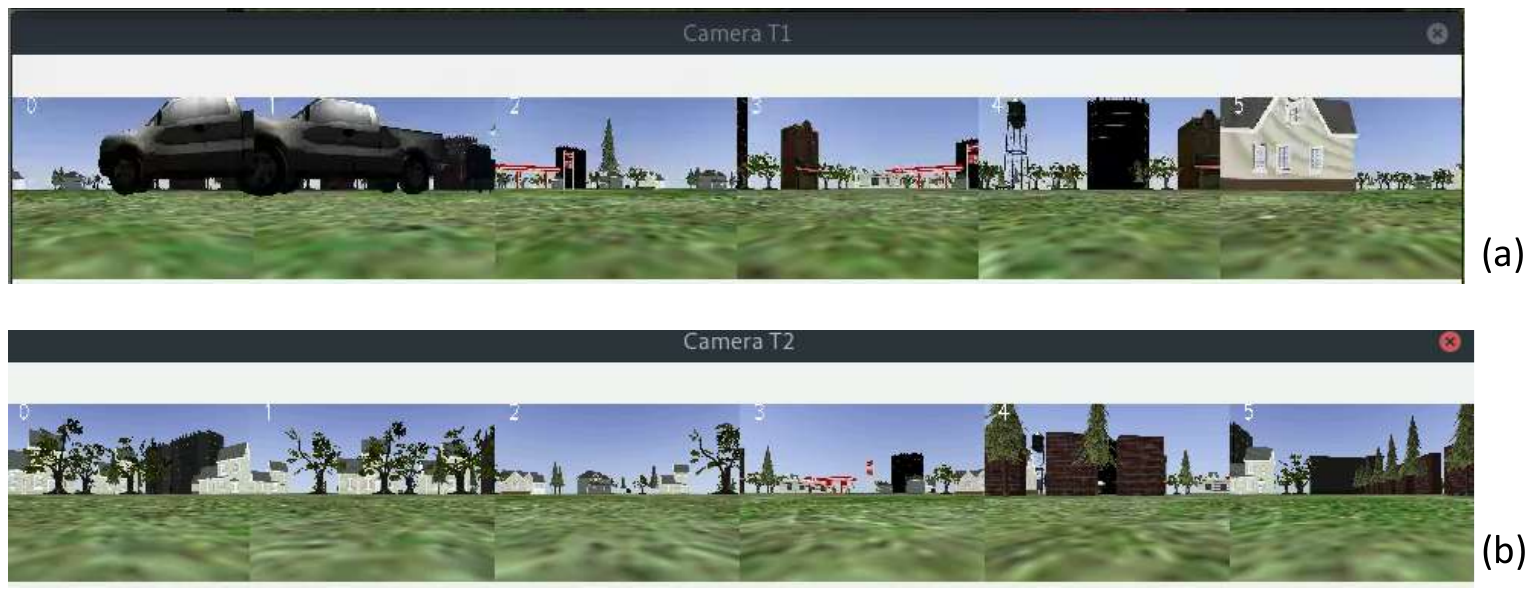}}
\vspace{-1in} 
\caption{The top figure depicts example robot pair positions showing a ROS/Gazebo 3D Urban landscape, and the bottom two show panoramic views from the robot cameras.}
\label{fig:eval1}
\end{figure}

One of the common and widely used landmark recognition software/testbeds is written in Robot Operating System (ROS), which is a popular open-source middleware. Gazebo, a 3D simulation engine, has also been integrated with ROS to provide robot software simulation testing. Further, in \cite{Lyons_2022} modified UCIC software was used in conjunction with two Pioneer P3AT robots equipped with cameras for this simulation. The modifications to the software include the blockchain and consensus implementation.

Fig. \ref{fig:eval1} depicts an example scene from our ROS/Gazebo suburban simulation. This simulation runs over a Digital Storm Aventum equipped with an Intel Core-i9 processor and GeForce RTX 3080 GPU and models a $130\times{}180 m^2$ flat suburban area filled with vehicles, grass, buildings, trees, and other objects. The robot team has simulated and executed using a model based on ROS/Gazebo, with Pioneer 3-AT robots operating in an environment where their FOV has both shared and occluded portions.

\subsection{Mission Success and Overhead}
To carry out a preliminary examination of the proposed BC-VH framework, we assessed the overhead associated with the end-to-end navigation (mission) in comparison with the traditional visual homing strategy (without blockchain). Here, we forced all robots to look around before the motion started rather than looking around as needed. We ran two missions with the blockchain feature on and two with it off, replicating the experiment twice with 8 missions in total.

\begin{figure}[htbp]
\centerline{\includegraphics[width=\columnwidth]{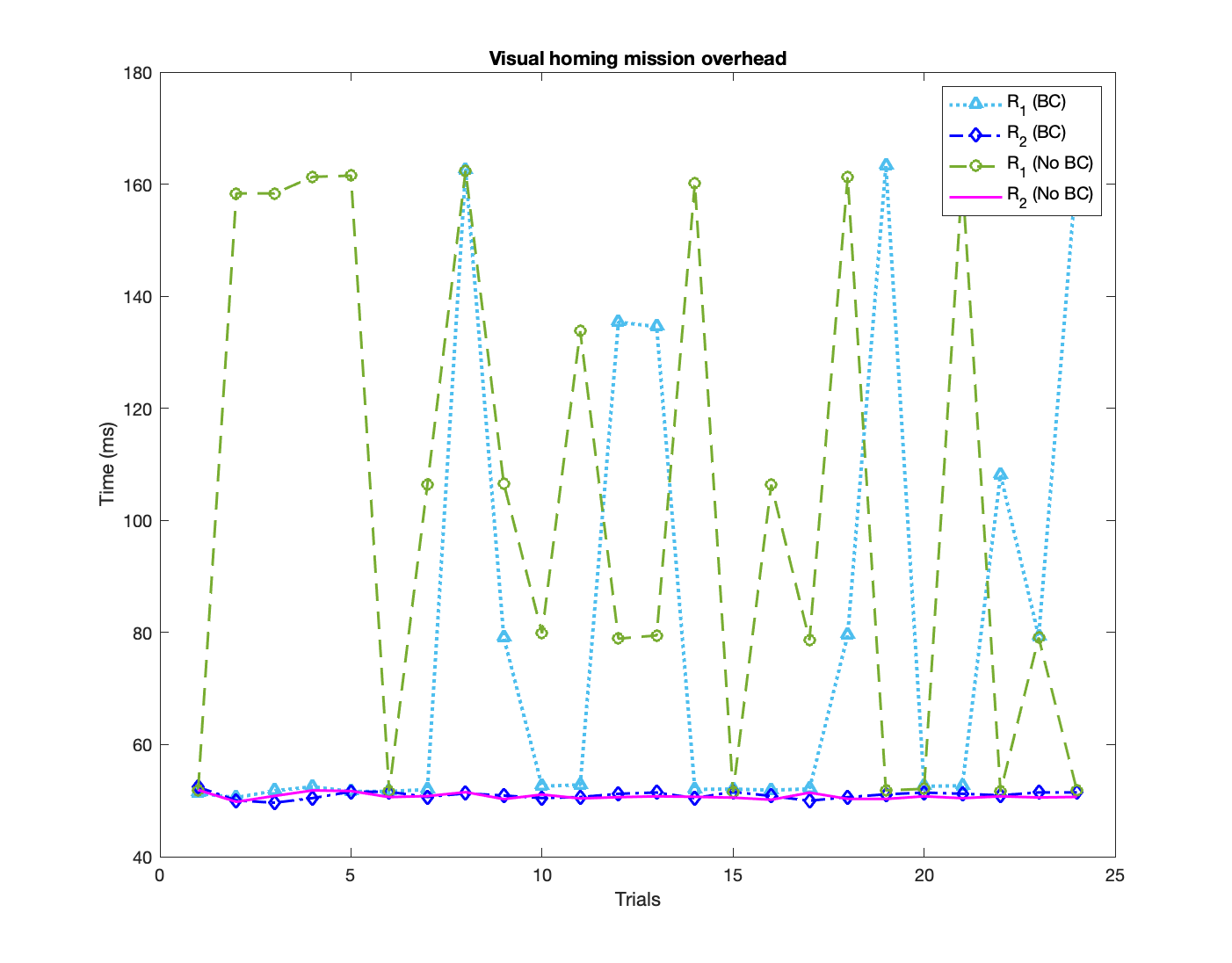}}
\caption{End-to-end delay in BC-VH vs. traditional visual homing strategy. \textbf{BC} and \textbf{No BC} indicate the usage of BC-VH and traditional/centralized visual homing (without blockchain), respectively.}
\label{fig:latency}
\end{figure}

Fig. \ref{fig:latency} depicts the overall time associated with mission completion by Robots $R_1$ and $R_2$. With blockchain implementation enabled, the robots can complete their mission with an end-to-end delay that is nearly equal to that of the blockchain-free visual homing system. That is, given all the benefits (i.e., tamper-proof data sharing, trustworthiness, decentralization, etc.) blockchain technology can bring into a distributed environment such as visual homing, it does not incur any significant computational delay in mission completions. The reason $R_1$ mission time is higher than that of $R_2$ could be due to the complex or cluttered environment and limited visibility conditions (e.g., obstacles). Table \ref{tab:latency} further presents the average delay/communication overhead associated with our blockchain solution over simulated FoVs with varying visual homing robot positions. The table demonstrates the significantly low overall delay associated with the substantial operations in the system, namely, ledger state update and panoramic view retrieval.

From the time complexity perspective, given $n$ evenly distributed robots (participating nodes) in the workspace:
\begin{itemize}
    \item Path planning: The maximum path length is $\ell\approx log_b(n+1)$, logarithmic in the size of the robot team. Assuming that each robot sees landmarks just in its vicinity, a worst-case path length would be $O(n)$.
    \item Blockchain operations: Multiple nodes collaborate to validate and record transactions. The time complexity of blockchain operations, such as transaction verification and block creation using PoW has a time complexity of $O(n)$, where $n$ is the computational difficulty set by the network.
    \item Communication and consensus: Assuming network propagation delay as nodes broadcast valid blocks to other nodes, the proposed PoW-based consensus is reached when the majority of nodes agree on the longest valid blockchain. The time complexity of communication and consensus in PoW networks is typically $O(n)$. However, the time complexity of PoW itself could be exponential in the worst case.
\end{itemize}

\subsection{Security Analysis}
Several security evaluation metrics are considered to investigate the effectiveness of the BC-VH framework employed for solving the cooperative navigation among the robots. We analyzed the effects of the following attacks in the BC-VH framework on the visual homing network performance. STRIDE \cite{ucedavelez2015risk}, a comprehensive threat modeling methodology was used to analyze and assess the strengths and limitations of the proposed prototype against a design of six categories.

\subsubsection{Spoofing}
Even the most diligent robotic nodes in the visual homing network could be tricked by sophisticated spoofing attacks. Here, the intruding robots can masquerade as a trusted source and illegitimately gain access to vital data shared by the authorized robots in the network. Spoofing generally takes advantage of the vulnerable visual homing networks, thereby collecting information for gaining network access controls.

\subsubsection{Tampering}
Sustainable maintenance of tamper-proof records is one of the inherent properties of the blockchain. When a block is created, the hash of the previous block will be added as a seal to the new block. In the BC-VH framework, in order to tamper with a block in the blockchain, the intruder has to tamper with every subsequent individual block in the chain associated with the robots in the network. Authenticated control over almost 51\% of the blocks in the BC-VH network is mandatory in order to recognize the tampered node, which is a challenging trend.

\subsubsection{Repudiation}
Repudiation provides proof of the occurrence of activity in the network made by a legitimate robot in the visual homing network. With repudiation, the transmitting robot will have proof of delivery, and the receiving robot in the network will have sufficient proof of the identity of the sender. Trustworthy issues are normally exploited with the repudiation, by denying the authenticated information shared by the robots to other legitimate nodes, which could be effectively addressed through the BC-VH framework.

\begin{table}[ht]
\caption{Average delay overhead in BC-VH over simulated FOVs with varying robot positions.}
\label{tab:latency}
\centering
\begin{tabular}{|p{1.3cm}|p{2.3cm}|p{2.4cm}|}
 \hline 
 \# Positions & Update Time (ms)  & Retrieval Time (ms) \\ 
 \hline
 40 & 0.82998 x$10^{-3}$ & 1.79 x$10^{-5}$ \\
 \hline
 200 & 0.100515 x$10^{-2}$ & 1.99 x$10^{-5}$ \\
 \hline
 500 & 1.09 x$10^{-3}$ & 1.98 x$10^{-5}$ \\ 
 \hline
\end{tabular}
\end{table}

\subsubsection{Information Disclosure}
Information disclosure or leakage occurs when a robot in the visual homing network intentionally reveals sensitive information to potential attackers. Through smart contracts and audibility deployment in the BC-VH framework, a certain degree of assurance could be ensured on information disclosures in the visual homing network.

\subsubsection{Denial of Service (DoS)}
In DoS attacks, the hacker disrupts the availability of data or authenticated robots in the visual homing network through reduction, connection closures, data destruction, or resource exhaustion. The BC-VH solution addresses this issue by offering incentive-based mitigation through reward mechanisms. Attacks on the PoW mechanism in BC-VH would require significant mining power to undermine the information exchange among the robots in the visual homing network.

\subsubsection{Elevation of Privilege}
Elevation of privilege refers to a vertical privilege escalation scheme where lower privileged users grant themselves permissions to higher privileges. In the visual homing network, this could be exploited by slave robots to escalate their privileges using reserved means and authentications meant for master robots. To mitigate this, smart contracts can be deployed in the BC-VH framework to detect and counter privilege elevation attacks through early detection and appropriate measures.

\section{Conclusion} \label{sec:conclusion}
This work introduces a novel framework to address limitations in robotic visual homing environments. The framework utilizes decentralized blockchain technology for lightweight navigation of a heterogeneous robot team in a wide, out-of-line-of-sight area. Challenges related to navigation tasks and FOV sharing are explored, and the BC-VH architectural design is proposed to reduce resource consumption. Use case scenarios, consensus mechanism establishment, and a security assessment framework are discussed to evaluate the efficiency and trustworthiness of BC-VH.


\end{document}